\documentclass[letterpaper]{article}
\usepackage{natbib,alifeconf}

\usepackage{amsmath}
\usepackage{graphicx}

\usepackage{color}
\usepackage{algorithm2e}
\usepackage{comment}

\usepackage{hyperref}

\title{Bootstrapping of memetic from genetic evolution via inter-agent selection pressures}
\author{Nicholas Guttenberg$^{1}$, Marek Rosa$^{1}$ \\
\mbox{}\\
$^1$GoodAI}

\begin{document}
\maketitle
\setlength{\parskip}{0pt}

\begin{abstract}
We create an artificial system of agents (attention-based neural networks) which selectively exchange messages with each-other in order to study the emergence of memetic evolution and how memetic evolutionary pressures interact with genetic evolution of the network weights. We observe that the ability of agents to exert selection pressures on each-other is essential for memetic evolution to bootstrap itself into a state which has both high-fidelity replication of memes, as well as continuing production of new memes over time. However, in this system there is very little interaction between this memetic 'ecology' and underlying tasks driving individual fitness --- the emergent meme layer appears to be neither helpful nor harmful to agents' ability to learn to solve tasks.

Sourcecode for these experiments is available at \url{https://github.com/GoodAI/memes}
\end{abstract}

Recent advances in machine learning have refined our understanding of ways that an individual agent might learn. Could we take advantage of ideas about social or cultural learning such as Dual Inheritance Theory \citep{cavalli1981cultural, boyd1988culture} in populations of agents to discover new kinds of learning algorithms? Whereas in an individual agent we might represent knowledge, skills, or behavioral norms as explicit memories, datasets, or neural network weights, in a social context information is kept alive via transmission from agent to agent. We investigate how message exchange in a population of (genetically evolving) agents can lead to self-replicating patterns, or 'memes' \citep{dawkins1976selfish}, which may be capable of non-genetic adaptation.

\begin{figure}[t]
\includegraphics[width=\columnwidth]{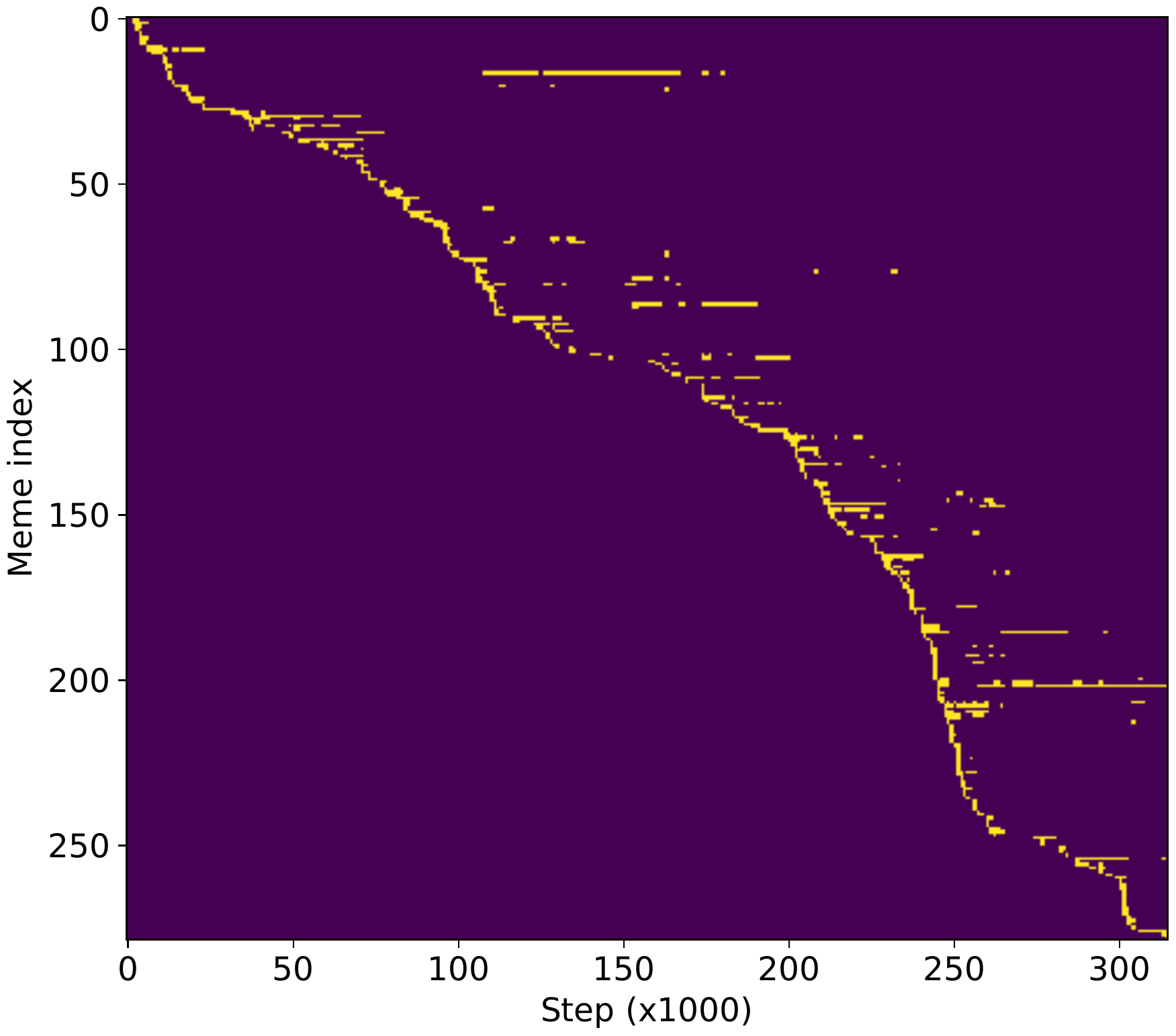}
\caption{\label{MemeEvolution}Evolution of memes over a $300000$ step run of our simulation. Unique messages are indexed on the vertical in order of appearance, with time on the horizontal. Bright pixels indicate that the message had a population over $80$ at that time.}
\end{figure}

Our motivation is to discover learning algorithms for multi-agent cognitive systems \citep{rosa2019badger} based on social and cultural adaptation. We pursue a 'Learning to Learn' style of approach \citep{hochreiter2001learning} in which the learning algorithm is discovered rather than designed. Here we focus on initial stages of this goal by trying to understand what it takes for communication dynamics to lead to adaptation in messages for the messages' own sake. The main idea is that while finding a message that replicates regardless of the underlying agent's network weights may be impossible in general, if the memetic degrees of freedom could influence the genetic evolution of the agents then memes could cultivate their environment for their own benefit by driving their hosts' evolution towards becoming better at propagating memes.

We implement this by allowing agents to exert selective pressures on each-others' replication on the basis of the degree to which an agent generated messages that others chose as salient. This can also be interpreted as agents having dynamic goals --- while an agent doesn't get to choose its own goals in this system, collectively the society of agents can choose different goals (here in the form of desired properties of messages) for each-other. In the end, this agent-provided selection pressure drives the continuing emergence of an array of new messages which are replicated with high fidelity (Fig.~\ref{MemeEvolution}). 

The continuing production of new messages alongside the survival of particular old ones is a surprisingly rich and apparently ecological process. Rather than a single 'most fit' meme dominating for all time, we see an ongoing exploration of the space of messages, something which may be a necessary condition (but not a sufficient one) to produce an open-ended learning system \citep{dolson2015s}. While specific 'static' sets of dominant messages emerge in our simulations without inter-agent selection pressures, this ongoing production of new dominant messages does not appear without that ability. 

Finally, as we are interested in using this sort of memetic evolution to help agents solve tasks, we look at what happens when we expand the networks to also solve an attached task (the Bipedal-Walker-v3 environment from OpenAI Gym \citep{brockman2016openai}) via individual-level selection pressures based on task fitness. We give the agents a common point of contact in the form of a shared global hidden state that is accessible both for the message processing parts of the agent and the task solving parts of the agent, so that the memetics and task policy parts of the agent can interact. However, as this connection is fairly limited, we perhaps unsurprisingly find that there is very little interference between task-based selection pressures and inter-agent memetic selection pressures --- the memes neither help nor hinder in the solution of the task, and memetic ecologies can still eventually emerge (though not as frequently) even under very strong selection on task performance. 

\section{Background Discussion}

This section comprises a short review of related literature on cultural evolution, memes, and artificial memetic evolution and social learning, suggesting some questions that are interesting to keep in mind as we explore this system. One theme with previous studies has been the question of the benefits or harms of having a fast cultural mode of adaptation in the presence of a slower genetic mode.

The concept of a meme as a self-replicating message, idea, or group of ideas was coined by Dawkins \citep{dawkins1976selfish},  around the time of the development of Dual Inheritance Theory \citep{cavalli1981cultural, boyd1988culture}. Dual inheritance is the idea that cultural and genetic adaptation can both occur, but are transmitted in related but not exactly equivalent ways, meaning that certain things spread through populations on the basis of being communicated or taught (oblique transmission), while others occur in line with heredity (vertical transmission). While all else being equal, genetic forms of adaptation may dominate over vertical memetic transmission \citep{cavalli1983cultural}, factors such as sexual selection for traits \citep{laland2008exploring} or strong oblique transmission can lead to the dominance of cultural modes of inheritance. 

The ideas of multilevel selection or group selection \citep{okasha2005multilevel} in biology suggest that another thing which might happen is that cultural and genetic adaptation may organize to respond to different selection pressures appropriate to their scale. In this context, the cultural degrees of freedom would act to collect and propagate group-level selection pressures down to the individual. These mechanisms can act in balance, or one may dominate, leading to (for example) the top-down control of replication in multicellulars. In memetic evolution, when would an (artificial) organism evolve to become a better host for its memes, versus when memes might evolve to improve the circumstances of their host organisms?

An example of memes directing group pressures down to the individual occurs in \citep{flentge2001modelling}, where memes were used to encode different social norms for agents and memetic transmission occured via a form of neighborhood voting. When this was used alongside a system with a suboptimal Nash equilibria with respect to individually-determined policies, the memetic degrees of freedom were able to capture selection pressures at the level of the group, in order to escape that suboptimal equilibrium and overall increase individual fitness against the underlying tendencies of the individual-level evolutionary dynamics.

We may also consider a process more like exaptation, in which the memetic dynamics end up being indirectly useful somehow to explore or organize a space of skills or abilities. In \citep{winfield2011embodied}, a system of mutually-imitating robots was able to discover policies for motion control simply through copying each-others' observed behaviors via imitation learning. Since some motions were easier to transmit than others, there was an implicit fitness associated with behaviors that could easily be passed on and executed consistently. However, there was no corresponding genetic degree of freedom in the system for the memetic evolution to compete with.

Another situation may be that the multiple levels of adaptation interfere with each-other --- rather than a conflict of interests or disagreement about selection pressure, fast memetic evolution may just make genetic evolution more difficult. This was studied in \citep{bull2000meme} by extending a the NKCS model \citep{kauffman1993origins} of a collection of independent but interacting replicators to capture a difference in timescale. Their results showed that with sufficiently strong epistatic interactions between genes and memes, fast memetic evolution (~10 times) could take over to the detriment of the slower genetic degrees of freedom, basically preventing the slow degrees of freedom from being able to stay adapted because of constant changes to their fitness landscape.

On the other hand, some systems appear to be able to use fast adaptation to effectively smooth the landscape for slower adaptation, actually making learning easier. In biology, this happens when organisms adapt in non-heritable ways within their own lifetime, something called the Baldwin effect \citep{baldwin1896new, mills2006crossing}. The within-lifetime adaptations mean that if there is a genetic change such as a limb becoming longer, rather than requiring a simultaneous genetic change to the way that limb is controlled and used (such that fitness only improved if both changes happened simultaneously), an adaptive nervous system can adjust that control policy on the fly, resulting in a less entangled and easier to traverse evolutionary landscape. 

When compared with the above NKCS model, the sorts of adaptations generally associated with the Baldwin effect are not heritable and therefore cannot compete with genetic replicators. Another aspect may be that the NKCS model has interactions which are fixed, random, and drawn from the same distribution --- in this case, adding more interactions between memes and genes automatically makes the landscape less modular and harder to evolve on. Its not clear that real landscapes would have this property --- some ways of decomposing function between memes and genes may align with natural modularities of tasks, leading to less overall entanglement between degrees of freedom. Meta-evolutionary pressures on such landscapes such as lineage selection \citep{virgo2017lineage} can encourage systems to find places where the evolutionary landscape is, for example, flatter \citep{wilke2001evolution} or where evolution is on the whole easier or faster. 

Given a more mechanistic model relating memes and genes, would the system remain stably in a state where memes are harmful to genes, or would they co-evolve to increase beneficial synergistic effects like the Baldwin effect? Furthermore, if the genetic degrees of freedom can work together with the memetic degrees of freedom, can we start with a system in which memetic evolution is low fidelity, random, and uncoupled from considerations of individual fitness and then have that system progressively evolve towards one in which social learning augments the individual-level learning capabilities? 

\section{Methods}

In this paper, we want to investigate how signals behave when transmitted and altered by a substrate formed of evolving neural networks, to see if we can characterize memetic evolution, genetic evolution, and their interplay. There are many different ways that this could be implemented in detail, so some of the architectural choices we made were motivated by desired features and properties, some determined during initial experimentation in order to find a region of the space of possibilities that gave rise to interesting dynamics, and some were basically arbitrary.

We sought a design in which agents would receive multiple messages and would be able to be selective about which ones they processed, as well as one which permitted the outgoing messages to be radically different from the incoming messages if that is what the network weights called for. We wanted there to be memory which would extend beyond a single message exchange so that individual agents could possibly propagate collections of messages, could become bored with messages that it had heard before, and could synthesize multiple incoming messages when determining what outgoing messages to produce. For practical reasons of being able to count distinct messages easily, we wanted the messages to be quantized rather than continuous.

To this end, we broadly chose to use a recurrent neural network for agents with a hidden state that would allow synthesis and combination over an extended interval of time. Incoming messages are stored in a memory buffer that contains exact copies of messages from the environment (keeping track of the sender). At each timestep, the agent's neural network generates a saliency associated with each message in its buffer, based on the agent hidden state and the contents of each message independently of the others. These saliencies are used with an entropy-adaptive softmax layer to attend over the messages, and to determine the agent's 'memetic input' during that step. The agent uses its current hidden state and current 'memetic input' to determine its next hidden state and to generate a message that it transmits to its neighborhood. These are the elements of the experiment that we thought were necessary to obtain the sorts of behaviors that would give rise to memetic evolution.

In addition to these elements, we found empirically that the presence of a skip connection from the memetic input to the memetic output has a strong effect on the diversity of memes that can propagate in a population made of random, identical networks. This can be thought of as a strong prior bias towards replication (because all else equal, the agent will tend to output a message that is similar to that which it hears from the environment). However, subsequent learning can erase this prior bias. The results we present primarily make use of this skip connection as it places the system in a richer region of the phase space, but we do include one experiment showing the effect of ablating the skip.

Beyond that, features such as the specific size of the network, the specific format of the messages, and the details of layers and architecture were more or less arbitrarily chosen. In these experiments we use messages that are themselves sequences, and use RNNs to both 'read in' the messages when determining saliency and updates, as well as to 'write out' the generated message. This has some potential benefits in the extensibility of the setup --- messages could be made shorter or longer without needing to change network weights --- but we don't think that this choice is strictly essential. For comparison, we will show a run using dense networks on vectorial memes rather than sequences. 

When we compare the effects of task-driven genetic fitness and emergent memetic fitness, we have to modify this architecture to include a network responsible for controlling the behavior of the agent during the task (in our experiments, this is BipedalWalker-v3). We add an additional RNN which takes as inputs the hidden state of the memetic RNN as well as the sensor inputs from BipedalWalker, and outputs actions and its own future hidden state. This RNN performs rollouts to completion during a single cycle of message exchange, so for a given instance of BipedalWalker the hidden state it receives from the memetic RNN is constant (e.g. the network only talks to other networks after the task is done). In our experiments, there was no feedback from the task-performing network to the message-generating network, something which we suspect limits the interaction between genetic and memetic evolution in our experiments and which we intend to change in future experiments.

\textbf{On computational costs}: We run these models on CPU, as due to each member of the grid only processing a small amount of data at a time and each grid site having different neural network parameters it runs about half as fast on GPU without particular optimizations. The individual runs we report took about a day each on a single machine, aside from our single 'long run' of $300000$ steps which ran concurrently with other runs for about six weeks.

\subsection{Network Details}

\textbf{Initialization}: In the baseline system, all weights of networks in the grid are initialized to the different values using orthogonal initialization with a gain of $4$ (which we refer to as 'heterogeneous initialization'). We also perform some experiments where the networks are initialized with the same weights (which is noted in those cases as 'homogeneous initialization').

\textbf{Messages}: The messages passed between agents are sequences of length 10, with three channels. Each channel at each position in the message can have a value of either -1 or +1. Agents are arranged on a grid, and receive messages other agents within a 5 by 5 square centered on themselves (24 messages per step). Each agent has a memory which stores up to 100 messages, with new messages pushing the oldest messages out of memory. A small amount of continuous Gaussian noise (standard deviation of $0.1$) is added to the messages as they enter memory --- this is something that made a larger difference in initial trial experiments where message generation by agents was deterministic and so noisy perception was a needed source of variation to observe mutation. However, in the experiments presented in this paper, message generation is inherently somewhat stochastic and so we expect this additional noise to be less important. We chose to retain it in order to keep results comparable between the experiments in this paper and some of the older ones that in the end we decided not to report in detail here. 

\textbf{Attention Network}: Each step, the Attention network generates a weighting over the 100 messages contained in memory. The Attention network is an RNN with a hidden state $\vec{h}_a$ of size 10. It takes as input the current 3-vector from the message being evaluated, the agent's current 'global' hidden state $\vec{h}_g$ of size 16, and its own current hidden state $\vec{h}_a$. During a step of the Attention network, it evaluates intermediate variables $\vec{G} = \sigma(L_G(\vec{x}))$ (the 'gate') and $\vec{U} = L_U(\vec{x})$ (the 'update') where $L_G$ and $L_U$ are linear layers with bias of size $29 \times 10$ and $\sigma$ is the sigmoid function. The next hidden state is then $\vec{h}_a^\prime = \vec{G} \vec{h}_a + (1-\vec{G}) \vec{U}$. A layer $L_L$ is used to project the final hidden state down to a single scalar, which is combined across all messages into a vector of attention logits.

\textbf{Adaptive Softmax}: Rather than directly taking the softmax of the attention logits, we use an adaptive method to adjust the logits so that the resultant attention distribution approximately obtains a desired entropy $H^*$. To do this, we apply softmax, calculate the current entropy $H$, and then rescale the logits by $1 + \alpha \frac{H - H^*}{H^*}$ with $\alpha=0.1$ in these experiments. We perform $20$ iterations of this rescaling. This way, if the entropy of the attention distribution is lower than the target, the logits are scaled to have smaller variation (raising the entropy of the softmax distribution), and vice versa. We choose a target entropy $H^* = 0.6$. This choice matters somewhat in that a very low entropy has a strong tendency to form single message centers that completely dominate the surrounding area, whereas with a high entropy the message contents doesn't matter. However, we do not present experiments on systematically varying this parameter in this paper, so our results may differ based on the choice of this value (or the choice to not fix the entropy of the attention over messages). Once the attention pattern has been determined, we resample from it using Gumbel-Softmax --- in the current experiments we present, this has no significant advantage over just sampling from the softmax distribution directly, but it leaves open the possibility of taking gradients of the overall process for future experiments. The result is that a single message $\vec{m}$ is sampled from memory.

\textbf{Agent Global Hidden State}: The agent global hidden state $\vec{h}_g$ is a vector of size $16$ which is updated once per message-passing step. There is a single layer $L_H$ taking the entirety of the chosen message as a vector of size $30$ plus the current value of the hidden state $\vec{h}_g$ as concatenated inputs $\vec{x}$. The next global hidden state is taken as $\vec{h}_g^\prime = \textrm{tanh}\left(\vec{h}_g + L_H(\vec{x})\right)$. Other architectural choices could have been to use an RNN to read out the message, use gated connections or an LSTM, etc --- we did not try these variations during development or experiment with them systematically, and initial experiments not using a global hidden state at all were still able to exhibit diverse replicating populations of memes. This element was added in order to enable message sequences over time or other cross-timestep structure to be possible if the course of evolution went that way, but for many of the observed results it is likely not essential.

\textbf{Message Generating Network}: The Message Generating Network has a similar architecture to the Attention Network, except that it outputs symbols in sequence. The Message Generating Network was implemented as a bidirectional RNN: it takes the attended message and its time reversal as inputs (sequences of 3-vectors), as well as its internal hidden state $\vec{h}_m$ of size 10 and the agent global hidden state $\vec{h}_g$. The choice of doing this as a bidirectional RNN was made before the introduction of the global hidden state, in order to include more non-local information about the message for the output. In retrospect, with the addition of the global hidden state it is probably unnecessary, but we retained it for comparison to older (not reported here) results.

It uses gated updates of the same form as the Attention Network. However, at each step, its internal hidden state is used via a layer $L_M$ to produce a 3-vector of logits. These logits are combined with the attended message to produce probabilities for each symbol of the output message to either be +1 or -1: $p(m^\prime_i = +) = \sigma( \beta (\vec{m}_i + L_M(\vec{h}_m)) )$. The constant $\beta$ is set to $3$, and was tuned to prevent messages from tending to decay towards full noise during the initial experiments. In experiments with deterministic generation of messages, this extra multiplicative constant had not been necessary, but on adaptation to stochastic generation we found that we could not get populations of non-random messages without it.

\textbf{Task Network}: The Task Network is an RNN that takes as input the Agent Global Hidden State $\vec{h}_g$ as well as its own hidden state $\vec{h}_t$ of size 16 and the 24 sensor inputs from the Bipedal Walker task. The network has two hidden layers of size 16 with ELU activations, after which there is a layer that produces the next hidden state using a $\textrm{tanh}(\vec{h}_t + f(x))$ activation similar to the global hidden state update, and a layer that produces 20-bin distributions of actions on all four action channels, from which the agent's actions are sampled. 

\textbf{Agent Replication and Evolution}: Agents in the system periodically replicate with mutation, replacing network weights at a neighboring site. Whether an agent replicates depends on two filters: one from inter-agent (memetic) selection pressures, the other from task-based selection pressures. Each step there is a chance per-agent ($0.1$ in the no-task experiments, increased to $0.2$ in the task experiments to see faster evolution) that it is chosen to 'promote' another agent in its neighborhood. This choice is made by weighting all neighbors according to how often the promoting agent selected that neighbor's message via the Attention Network: an agent is most likely to promote the source of messages it found salient. Once an agent has been promoted, if the mean fitness of the agent on the task is within the top $N$ agents ($16$ for these experiments), the agent replicates into a random neighboring site. A fraction ($0.001$) of offspring's weights are mutated as follows: the weight is multiplied by $0.99$ (weight decay) and Gaussian noise with standard deviation $0.2$ is added. The mutation fraction was adjusted during initial experiments to allow for stable behaviors to form --- if significantly higher, we observed that genetic drift would outpace the ability for stable replicating memes to emerge.

To modulate the strength of inter-agent and task-based selection pressures, we use parameters $\gamma_s$ and $\gamma_f$ respectively. A fraction of the time corresponding to $\gamma_s$ we pick a uniform random agent to promote rather than using weighted choice. A fraction of the time corresponding to $\gamma_f$, we ignore the fitness criterion and just allow the agent to replicate regardless.

\textbf{Bipedal Walker modified reward}: The BipedalWalker task has a strong negative contribution to the reward when the agent falls over, which tends to make samples for evaluating fitness very noisy. We use a variation where we fix the maximum length of the rollout to 400 steps and only count the best intermediate reward value obtained over the course of the rollout. This has smoother improvements than the original reward scheme, which in turn maintains a more consistent selection pressure that should provide a better measure of the balance between memetic selection and task selection over extended periods of time.

\section{Results}

\subsection{Memetic Selection Experiment}

First, we would like to establish the basic behavior of messages in this system. Do messages end up being copied and replicate, will networks generally output a message independent of their inputs, or will there just be chaos or random messages without much repetition? Furthermore, what basic elements of the setup are necessary for these phenomena? For this, we consider the case in which there is no external task or associated task fitness, only meme-driven selection effects ($\gamma_f = 1$). We count each unique message and see which messages occur with high frequency, and if and when new messages appear. From this point we will use the term 'meme' with regards to the specific messages in our simulation as a shorthand to refer to messages which have a higher population than would be expected from random generation, though we recognize that the term has other implications which these high-population messages may not exhibit in every case. 

\begin{figure}
\includegraphics[width=\columnwidth]{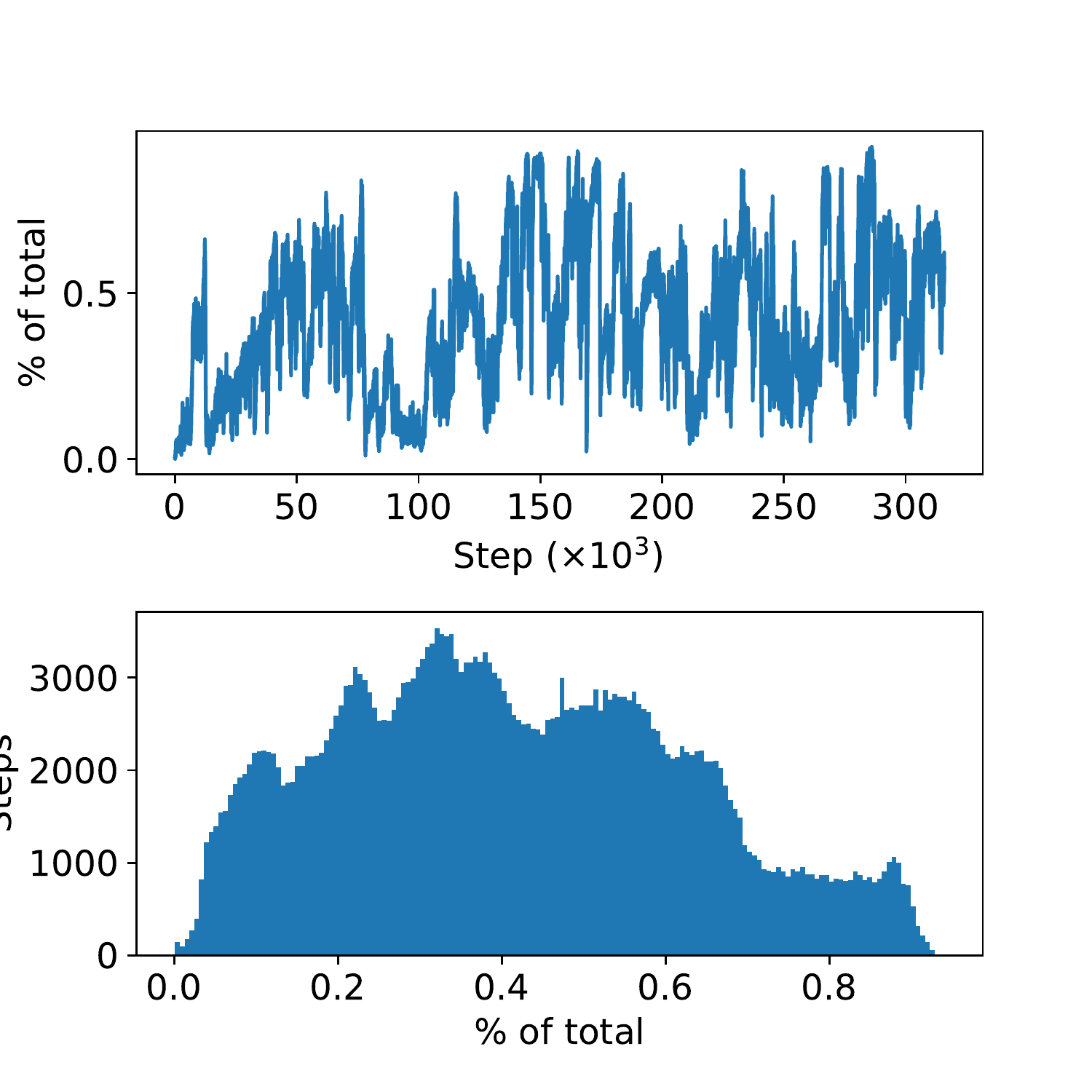}
\caption{\label{MemeDominance}Population of the most populous meme in the long run, versus time (top) and as a distribution (bottom).}
\end{figure}

When we run the baseline setup described in the Methods section on a $32 \times 32$ grid, within about one thousand steps we find that repeated messages with copy numbers in the population greater than $40$ reliably emerge. In our longest run of around $300000$ steps, we observed periods in which a small number of messages dominated with a maximum observed coverage of around $94\%$, alongside other periods in which the most populous messages had $20\%$ coverage. In Fig.~\ref{MemeDominance} we show the time-series of the population of the most populous message over the course of the long run, as well as a histogram of the coverage fraction of the most populous message when we divide time into windows of 1000 steps each. 

We plot the general course of evolution of the suite of messages from our long run in Fig.~\ref{MemeEvolution}. Overall, rather than seeing the same messages come and go, we see a continual emergence of new messages as the run progresses. This emergence of new messages is roughly linear with time over the long run, though it has fast periods and slow periods which suggest local stagnation and disruptive events in the evolutionary dynamics. One interesting feature is that some messages are stable for very long periods of time (~50000 steps) even when other messages are rapidly appearing and going extinct alongside (but still with populations comprising on the order of $10\%$ of the available space). Even when one message makes up a significant proportion of the population, there are generally on the order of $10$ distinct messages with at least $10$ copies, and on the order of $5$ messages with at least $20$ copies in the system at any given time. This coexistence between churn and stability seems more reminiscent of ecological dynamics (e.g. systems where the effective 'fitness' of a lineage is driven more by interactions between heterogeneous things) than of competition between things within a niche.

Given that we do observe messages with high copy number, we think it fair to conclude that something in the system is being copied. However, it could be that what we're seeing is not messages being copied because of their adaptation to hosts, but rather host genetics spreading as agents in the underlying grid just passively replicate and the genetic diversity of the population drops. In order to check this, we measure the number of distinct messages that exhibit populations above $40$ and the highest population measured during runs of the following ablations (runs of length $10000$, but still on the $32 \times 32$ grid):

\textbf{No evolution}: In this experiment, we randomly initialize networks separately but deactivate the evolutionary dynamics, so that each grid site has a fixed network with random weights. This tests to see whether memes can spread on their own merits (in a network-independent fashion). We find no high-population messages, suggesting that message contents alone cannot overcome the difference between network weights --- for messages to spread or become common, they must be co-adapted to the network which spreads them.

\textbf{No variation}: In this experiment, we fix the networks in the system to all have the same weights and turn off mutation. This effectively disables all genetic modes of operation in the system --- if memes arise or diversify, it would be entirely due to interaction with the shared, randomly chosen network weights. In this case we see some initial diversity of memes, but rather than that diversity growing with time it saturates.

\textbf{No mutation}: Here we want to test if the diversity of memes arises primarily from genetic diversity. We initialize the networks independently and turn off mutation, but leave replication and memetic selection in place (so that networks initialized with 'good' weights can still dominate, but the diversity of networks can't increase during the run). Without mutation, even given initial diversity of weights, we do not observe the emergence of messages with large populations.

\textbf{No skip connection}: We turn off the skip connection in the Message Network, to see if it is basically making message replication the default behavior of the system, or whether diverse high-count messages can emerge even without an explicit bias towards outputting the message that is attended to. The results indicate that large populations of messages can occur without the skip connection, but we do not see the emergence of a diverse set of high-population messages, suggesting that biasing the system towards a default 'copy-like' behavior may be important for obtaining an ecosystem of memes rather than a few dominant motifs.

\textbf{No selection + heterogeneous networks}: Here we replace the 'promotion' mechanism with a random choice of neighbor, rather than basing it on the fraction of messages that were attended to by neighbors. We retain mutation, but initialize the networks with different weights, so that there is a large degree of initial variation. In this case we see much smaller populations of messages overall, suggesting that replication or spread of particular messages is much more difficult without the mechanism of memetic selection. This confirms that memetic selection in this system can drive the genetics to be beneficial to the spread of memes.

\textbf{No selection + homogeneous networks}: Here we replace the 'promotion' mechanism with a random choice of neighbor, rather than basing it on the fraction of messages that were attended to by neighbors. We retain mutation and initialize all networks to the same weights. Despite the homogeneous network weights, without memetic influence on selection we do not see the spread of populous memes.

\textbf{Simplified Messages}: Instead of a 3-channel message of length 10, we change this to a 30-channel message of length 1 effectively replacing the recurrent sequence-processing parts of the agents and messages with dense networks applied to flat vectors. While we see some reduction in the diversity of high-population messages generated over the course of the run relative to baseline results, the highest populations achieved are similar. This suggests that the particular choice of using message sequences over vectors or other forms of messages is not critical to our results.

The results of these experiments are show in table \ref{ExperimentResults}. We include results from multiple runs separately where we have them available.

\begin{table}
 \begin{tabular}{cccccc}
  Experiment & Max & $\#$ & Mut. & Same & Sel. \\
  & pop. & $>40$ & & Init & \\
  \hline
  No evolution & 6 & 0 & N & N & N \\
  \hline
  No variation & 153 & 9 & N & Y & N \\
  \hline
  No mutation & 25 & 0 & N & N & Y \\
  \hline
  No skip & 454 & 4 & Y & Y & Y  \\
  \hline
  No selection/Hom.  & 12 & 0 & Y & Y & N \\
  \hline
  No selection/Het. & 81 & 1 & Y & N & N \\
  No selection/Het. & 21 & 0 & Y & N & N \\
  \hline
  Simplified & 502 & 11 & Y & N & Y \\
  \hline
  Baseline & 496 & 43 & Y & N & Y \\
  Baseline & 446 & 16 & Y & N & Y \\
  Baseline & 453 & 21 & Y & N & Y \\  

 \end{tabular}
 \caption{\label{ExperimentResults} Results of ablation experiments. Horizontal lines separate different experiments, and multiple entries within a block show different runs using the same protocol. Max pop. indicates the highest observed instantaneous population of any one individual message.}
\end{table}

\subsection{Interaction Between Fitness and Memetic Selection}

\begin{figure}
\includegraphics[width=\columnwidth]{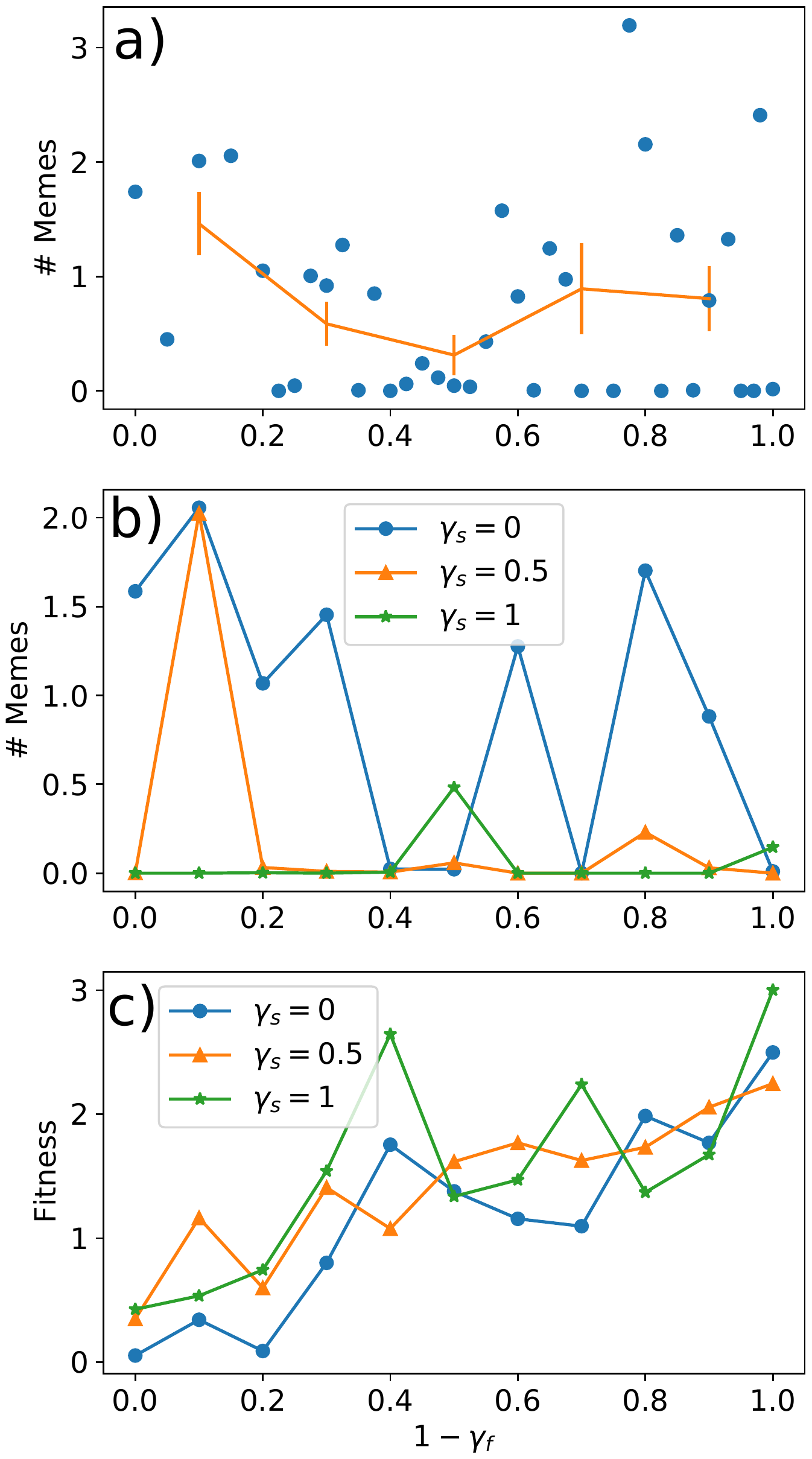}
\caption{\label{SelectionEffects}a) Runs where $\gamma_s = 0$ (maximum memetic selection strength) but in which $\gamma_f$ is varied. The solid line shows the average number of memes in bins of size $0.2$ in fitness selection strength. b) Points for $\gamma_s = 0$, $\gamma_s = 0.5$, and $\gamma_s = 1.0$, as a function of $\gamma_f$. Each point is an individual run. c) Effect of both $\gamma_s$ and $\gamma_f$ on task fitness. Each point is an individual run.}
\end{figure}

If we have a system in which both memetic selection and individual-level fitness-driven selection are taking place, do the different pressures interfere, or do memetic degrees of freedom evolve into a symbiotic relationship with individual-level fitness? To test this, we performed a number of runs of a $16 \times 16$ grid of agents performing the BipedalWalker task alongside the message passing dynamics for a period of 1000 steps. We vary both $\gamma_f$ and $\gamma_s$ to see the effects of different selection pressures on the emergence of messages with significant population counts. We show the number of memes with populations $8$ or more copies as a function of the task selection strength ($1-\gamma_f$) in Fig.~\ref{SelectionEffects}a, and the dependence on memetic selection strength ($1-\gamma_s$) in Fig.~\ref{SelectionEffects}b.

There seems to be no value of $\gamma_f$ where memes with relatively large populations cannot emerge, but rather it seems that emergence of memes is bimodal. In some runs, one or more high-population messages emerge, whereas in other runs there are none observed. It seems as though strong selection for task fitness can somewhat suppress or delay the emergence of memes, but this may be due to an overall lower level of successful replication events (since the task fitness selection filter rejects replication events that would have gone through in the $\gamma_f = 1$ case). On the other hand, we see a very strong effect with $\gamma_s$, where reducing the ability for message attention to influence genetics strongly reduces the rate at which memes appear.

We can also look at the effect of these terms on fitness. Fig.~\ref{SelectionEffects}c shows the fitness as a function of both $1-\gamma_f$ and $1-\gamma_s$. While changing selection pressure for task fitness has a strong (and roughly linear) effect on the ultimate task fitness obtained by the system, there seems to be very little effect on task fitness from memetic pressures, and so memetic influence on genetic changes seems to be independent of improvement on the external task --- it neither helps nor hinders. Part of this is likely because there are no particular aspects of this task and architecture where the memetic degrees of freedom could significantly assist things (more on this in the conclusions), but it may also be that the degrees of freedom that the memetic considerations exert pressures on are sufficiently disentangled from task degrees of freedom that the two selection pressures more or less interleave without impacting each-other.

\section{Conclusions}

Our results seem to indicate that by allowing agents in a system to exert selective pressures on each-other based on exchanged messages, there is a tendency for the mechanisms of attention and message transmission to evolve towards supporting a population of not only self-replicating, but also diversifying memes. One striking thing was how important that mechanism for agents exerting selection pressures on each-other was --- while having a population of identical networks was sufficient to obtain a small number of high-population memes, we only saw a high diversity of such memes when agent preferences for messages translated into replicative pressures.

The diversity of observed memes in such cases is interesting from the perspective of open-endedness. Normally if we had a system of replicators that each had fitnesses independent from one-another we would expect to see competition leading to dominance of the highest fitness quasispecies and a general collapse of diversity.  Interestingly, the memetic replication that emerges from our system does not seem to go to that end-point. Rather, while there are a subset of messages which comprise up to $90\%$ of the population, those dominant messages are repeatedly overtaken by new memes in a process that seems more reminiscent of ecological cascades and turn-overs than single-niche dynamics. Systems with a single shared fitness function, or even ones which converge towards a point of stability (optimizing some associated Lyapunov function) are 'homeostatic' and in some sense are ultimated closed, whereas systems that are able to change what they are trying to accomplish at any given point can be 'heterostatic' \citep{oudeyer2009intrinsic}. While this is not a sufficient condition for systems to exhibit coherent exploration or continual creation of novel functions (rather than just novel forms), it is likely to be at least a necessary condition. So this is a property we want in some putative social learning algorithm, and the fact that we didn't have to go out of our way to obtain it is nice.

Unfortunately for the goal of using these memetic dynamics to discover novel learning algorithms, we found little interaction between task-driven fitness and the memetic dynamics. Messages passed between agents seemed to neither help nor hinder the discover of better policies for the Bipedal Walker task. We identify three factors to investigate for future work in order to try to see whether the meme layer can give rise to more interesting learning dynamics.

One factor is that there was no feedback from the details of performing the task back into the messages --- agents could not in any sense 'report' on their success or failure, on what happened, on what they tried, etc. While the existence of such a feedback would not necessarily guarantee interaction between the memes and the task, the absence of that feedback prevents the memetic selection pressure mechanism from exerting pressures aligned with task performance or behavior on the task.

A second factor is that the sorts of things that a rapid, communication-driven learning method is less useful when the task is static, because information that is communicated on personal timescales becomes redundant on evolutionary timescales. Tasks whose switching timescale is similar to the evolutionary timescales of the memes themselves might allow for benefits which could not have been realized in our system. That is to say, if rather than a single task we had different variations of the task which were correlated in space and time and which did not easily admit a single policy that would generalize over all variants, memes could relay locally relevant information about changes to the task before the genes had a chance to catch up, basically implementing something similar to the Baldwin effect.

A third factor is that our task was entirely individual and did not explicitly involve any group behavior or coordination between agents. The potential benefit of memes as norms or even as shibboleths identifying groups likely to cooperate or defect could not be realized in our system. Something like a population-level Iterated Prisoner's Dilemma might be a more appropriate task to see interactions between memetic and genetic evolution.

\section{Acknowledgements}

We would like to thank Jan Feyereisl, Isabeau Premont-Schwarz, Lisa Soros, and Olaf Witkowski for feedback and suggestions for refining this paper. 

\footnotesize

\bibliographystyle{apalike}
\bibliography{references}

\end{document}